# Adaptable, shape-conforming robotic endoscope

Jiayang Du[1], Lin Cao[1], Sanja Dogramazi [1]

[1]School of Electrical and Electronic Engineering, The University of Sheffield,

Sheffield, United kingdom

**\*Correspondence:**
Jiayang Du
DDu4@sheffield.ac.uk

**Abstract**

This paper introduces a size-adaptable robotic endoscope design, which aims to improve the efficiency and comfort of colonoscopy. The robotic endoscope proposed in this paper combines the expansion mechanism and the external drive system, which can adjust the shape according to the different pipe diameters, thus improving the stability and propulsion force during propulsion. As an actuator in the expansion mechanism, flexible bellows can provide a normal force of 3.89 N and an axial deformation of nearly 10mm at the maximum pressure, with a 53% expansion rate in the size of expandable tip. In the test of the locomotion performance of the prototype, we obtained the relationship with the propelling of the prototype by changing the friction coefficient of the pipe and the motor angular velocity. In the experiment with artificial bowel tissues, the prototype can generate a propelling force of 2.83 N, and the maximum linear speed is 29.29 m/s in average, and could produce effective propulsion when it passes through different pipe sizes. The results show that the prototype can realize the ability of shape adaptation in order to obtain more propulsion. The relationship between propelling force and traction force, structural optimization and miniaturization still need further exploration.

## 1    Introduction

The incidence of colon diseases is rising day by day, which is closely related to modern people's lives and eating habits, such as sedentary, low dietary fiber intake, mental stress [1], etc., which all can lead to colon diseases, including colon cancer [2]. Colonoscopy, as the gold standard for colon examination, provides a visual examination of the colon lumen and effective treatment. Studies have found that regular screening is an effective approach to preventing colon diseases, which can effectively reduce the incidence of colon cancer [3].

As a reliable means of intestinal examination, colonoscopy has been used by countless patients. In the procedure of colonoscopy, the clinician inserts the colonoscope using one hand to control insertion length, speed, and force of colonoscope propulsion, and the other hand to control the direction of the semi-flexible tip of the colonoscope by controlling the two dials. The acceptability of colonoscopy by the patient remains a significant issue [4], since the procedure has hazards of colonoscope looping, intestinal extensions, discomfort, and perforation, which can cause injury to the patient both physically and psychologically [5]. Moreover, colonoscopy clinical training is rigorous and often exhausting, which can result in repeated strain injuries to the hand and wrist muscles and ligaments. All of these have served as compelling incentives for researchers studying robotic colonoscopy to find a more practical and pleasant way to use the current technology.

To alleviate the challenges associated with conventional colonoscopy for both patients and clinicians, as well as to enhance the accessibility of colonoscopy, researchers have started exploring more intelligent and automated alternatives.

Many robotic platforms for colonoscopy have been proposed in the last couple of decades. Since the introduction of the miniature ingestible capsules with a camera for recording the colon internal wall in 2000, the focus has been on their controlled propulsion (rather than peristalsis [6], [7]) for detailed screening of colon [8], [9]. Gregory and others designed endoscope equipment that drives the crawler to move through the meshing of worms and gear, but the movement efficiency is still an urgent problem [10]. About the influence of contact area on the propulsion of robotic colonoscope, some studies have shown that for robots that rely on crawling, or wheel movement, the contact area of the abdomen is usually limited, which leads to insufficient friction and traction and affects the stable movement [11] [12]. The ball bearing adopts a wheel structure. This unique symmetrical wheel design allows it to fit closely to the intestinal wall, increasing the contact area and improving traction, allowing it to pass through narrow intestines[13]. This method is also favoured by researchers in the field of pipeline robots [14]. Similarly, Vanni et al. put forward a robot endoscope based on wall pressure, which uses two independent balloons and a crawler transmission system to increase the diameter of the robot by inflating the balloons, enhance the contact between the crawler and the intestinal wall, and ensure stable traction [15].

Several attempts have been made to drive a capsule with an inserted magnetic coil from a set of external magnets [16], [17], [18] but the complexity of the human tortuous colon and friction forces between the capsule and the colon often prevent complete success of this interesting approach. Robotic colonoscopes have also been inspired by inchworm, snake and centipede locomotion to propagate through the colon with effectors that require multiple continuous contact with the colon wall for anchoring, traction or footholds. The anchoring aspect in the colon soft cavity accompanied by mucus is still a problem [7] while gaining sufficient traction and foothold to push the robot forward [1], [8], [9] can result in tissue damage. A recent work on traction based robot mechanism which consists of two independent balloons and a transmission system based on tracked motion [15] has been reported to successfully navigate the colon lumen. The wall-pressing balloons maintain the endoscope in the central position of the lumen and obtain traction through several tracks that contact the colon wall. Even though the force that propels this tethered head provides tip force, its ability to steer around sharp colon angles is not realized. Furthermore, the traction force cannot be maintained throughout the colon that has varying diameter.

In this paper, we present a new robotic endoscope prototype that takes inspiration from [15] and the author's earlier work in this area [19] to advance the expansion mechanism feature to overcome the shortcoming of radial adaptation throughout the colon length. An external actuation with worm-gear coupling to an expansion mechanism based on flexible bellows control extendable scissor mechanism that maintain pressure on four sets of flexible tracks. This design allows the robot to adapt to varying colon sizes while maintaining stable propulsion and consistent contact with the colon lumen. The combination of external motor control and adaptive sizing improves locomotion by allowing the robot to adjust to varying colon sizes, ensuring stable contact and better traction. Furthermore, this design has the potential to steer the whole mechanism to negotiate sharp bends of the colon although this paper reports our preliminary work on the straight colon section.

## 2    Prototype Design

The human colon is a continuous hollow tube approximately 1.5 m long and 7.5 cm in diameter when inflated [8]. The human large intestine consists of six segments: rectum, sigmoid colon, descending colon, transverse colon, ascending colon, and cecum, with average diameters between 2.6 and 4.5cm; see table I [20].

Table I Average diameter of colon segments

| Colon sections | Average Diameter (mm) |
| --- | --- |
| Rectum | 3.6 |
| Sigmoid colon | 2.6 |
| Descending colon | 3.3 |
| Transverse colon | 3.7 |
| Ascending colon | 4.5 |
| Cecum | 4.4 |

In addition to dimensional changes, the human colon has certain elasticity, which enables it to effectively transport and treat food residues during digestion. It consists of multiple tissue layers, such as the inner mucosa, muscular layer, and outer serosa, which provide the colon with the strength and flexibility needed to expand and contract under pressure. The diameter of the colon varies significantly along its length, reflecting its different functions. For instance, the larger diameters of the ascending colon and cecum accommodate water and electrolyte absorption, requiring more space for storing and processing contents. In contrast, the narrower sigmoid colon and rectum are primarily responsible for storing and expelling faeces and thus do not need as much capacity as the ascending sections.

### 2.1 Design

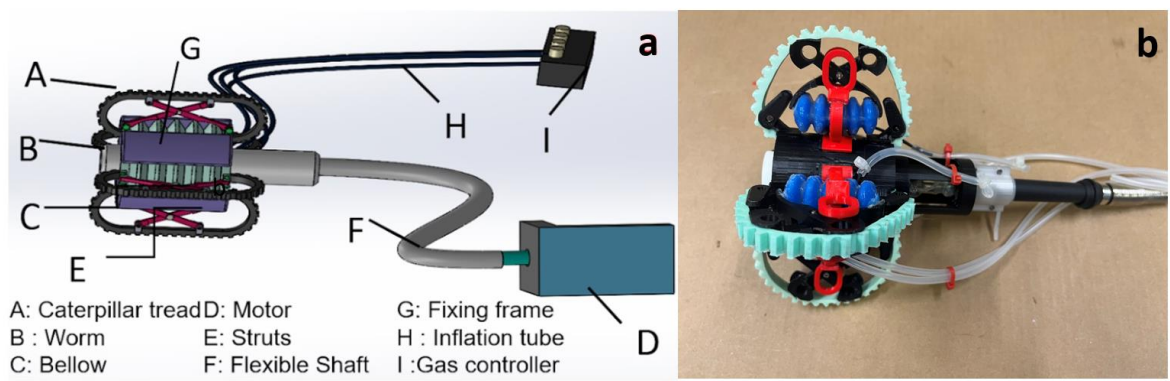

Fig. 1 a) A conceptual diagram of the robotic endoscope. The endoscope tip is driven using a worm couple (B) actuated by an external motor (D) through a flexible shaft. The tip also features four independently controlled expansion mechanisms enabled by a pneumatic bellow (C) and strut mechanisms (E), b) The expandable tip. The external motor is connected to a Arduino UNO to realize locomotion of the device (not equipped with camera lens).

The structure of the flexible shaft, which is composed of a metal shaft and an external rubber sleeve, is shown in Fig. 1. Flexible shaft is widely used on occasions that need to transmit rotary motion but require a certain degree of flexibility. However, it cannot be ignored that the interaction between the metal shaft and the rubber sleeve leads to frictional contact during the rotating movement, which leads to power dissipation. The flexible bellows are separately controlled, and their displacement adjusts the scissor-shaped rigid struts to maintain contact between the tracks and the inner wall of the colon. The inflation of bellows and consequent extension of the struts determine the normal force exerted by tracks on the inner wall of the colon. The smooth back surface of each track slides over the groove reserved in the strut. On the other side, the external toothed surface of the track contacts the colon wall and generates friction, which propels the whole structure through the colon.

### 2.2 Modeling

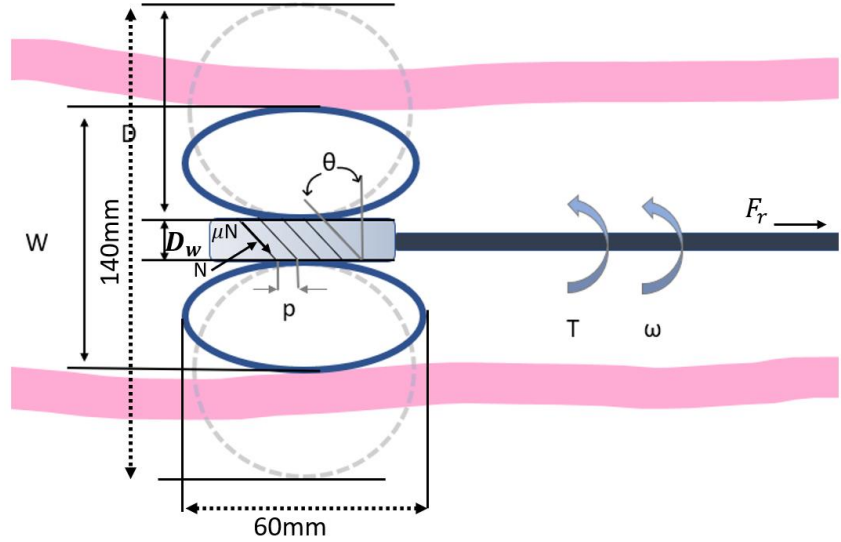

Fig. 2. Diagram of the kinematic mechanism of the expansion tip.

Inspired by [8], we describe the motion of the robot in the colon, as shown in Figure 2. We made a static analysis in the colon lumen, the tracks deform to the colon diameter and exert pressure on the colon wall. In the simplified model, the colon is considered as a rigid tube, which does not deform, therefore, it is more convenient to analyze the force of the tip of the robotic colonoscope in the colon, and more intuitive to show the working principle of the device in locomotion and expansion mechanism.

In this case, the maximum static friction between the colon wall and the track is:

$$F_f = n\mu_c N = n\mu_c k(D - \frac{W-d}{2}) \qquad (1)$$

Where n denotes the number of tracks, $\mu_c$ is the frictional coefficient between the track and lumen wall, k is the radial equivalent elastic modulus of track, D is the diameter of track, W is the diameter of colon lumen, d is the diameter of the worm gear.

The motor drives the only worm, and the angular velocity of the motor can be expressed as:

$$v = \frac{d\tan\theta\omega}{2} = \frac{\omega p}{2\pi p} \tag{2}$$

Where $\omega$ is the angular velocity of the motor, p is the pitch of the worm, v is the velocity of the robot.

Thrust force and torque from the worm can be presented, shown in Fig 2.

$$F_r = N\cos\theta - \mu N\sin\theta \tag{3}$$

$$T = \frac{d}{2}(N\sin\theta - \mu N\cos\theta) + T_f \tag{4}$$

Where N indicate the normal force on worm, $\mu$ is the frictional coefficient between the worm and the track, T is the output torque of motor, $T_f$ is the reaction torque (friction between worm and tracks).

According to the equation Pitch angle of worm,

$$\tan\theta = p/d\pi \tag{5}$$

Therefore, from prior Eqs. (3) (4) and (5), the relation between motor torque and thrust force can be derived as follows:

$$F_r = \frac{2(\pi d - \mu p)(T - T_f)}{d(\mu\pi d + p)} \tag{6}$$

if we assume the relation between output torque and rotation speed is linear, it can be determined as follows:

$$\omega = \omega_n(1 - T/T_M) \tag{7}$$

Where $\omega_n$ is angular velocity under no load, $T_M$ is maximum torque when angular velocity is zero.

Therefore, according to Eqs. (6) (7), the max thrust force of the robot can be determined as follows:

$$F_{MAX} = \frac{2(\pi d - \mu p)(T_M - T_f)}{d(\mu\pi d + p)} \tag{8}$$

The transmission ratio as follows:

$$i = \frac{Z_w}{Z} \tag{9}$$

Where $Z_w$ and $Z$ represent the number of teeth of worm and track respectively, which are shown in Table 2.

The linear velocity of the worm can be calculated as:

$$v_w = \frac{\pi D n_w}{60} \tag{10}$$

Where $n_w$ is the rotation speed of the worm, we regard the rotation speed of the worm as the rotation speed of the motor without considering the sliding of the flexible shaft and the rational friction resistance. Finally, we get the speed of the prototype as follows:

$$v = i \cdot v_w \tag{11}$$

### 2.3 Structure and locomotion

The cylindrical shape of the robot is created by an assembly of four flexible tracks and four expandable bellows. The robot structure is divided into: 1. the power transmission system (a worm connected by a flexible shaft is engaged with the track), 2. expansion tip, an expansion mechanism with the four bellows and four struts, 3. the four flexible tracks in direct contact with the colon wall, and 4. the motor and its control system. The external motor transmits power to the worm gear via a flexible shaft, which, in turn, drives the flexible tracks to produce linear motion for the entire structure. Four tracks and four flexible bellows are symmetrically positioned around a cylindrical frame, with each track wrapped around the frame and joined at the ends to form a closed loop. The expansion mechanism adjusts the robot's external radius to maintain traction against the colon wall by extending the bellows, which changes the angle of the struts to push the tracks outward. This symmetrical structure allows a single motor and worm to drive all four tracks, making it well-suited for navigating narrow spaces with varying radial dimensions.

### 2.4 Prototype fabrication

Table2. Specifications of the fabricated robot.

| Parameter | Dimension |
| --- | --- |
| Worm teeth number, $Z$ | 5 |
| Track Teeth number, $Z_w$ | 34 |
| The prototype frame width (mm) | 140 |
| Track width (mm) | 7.5 |
| Belt height (mm) | 4.5 |
| Tooth length (mm) | 3 |
| Worm gear pitch (mm), p | 3 |
| Worm gear diameter (mm), $D_w$ | 28 |

The actuation of the prototype is provided by an external stepping motor (NEMA 17, Rtelligent) which is controlled by a microcontroller (Arduino Uno). The rigid worm gear and outer sleeve are using PLA obtained by 3D printing. The innermost sleeve has four grooves that limit the relative sliding and lateral movements of the track. The outermost frame covers the worm and tracks. In order to avoid excessive friction between the tracks and the sleeve, the track groove is shaped as an arc and filled with lubricant. Bellows are created in two parts and bonded together. (see Fig 3).

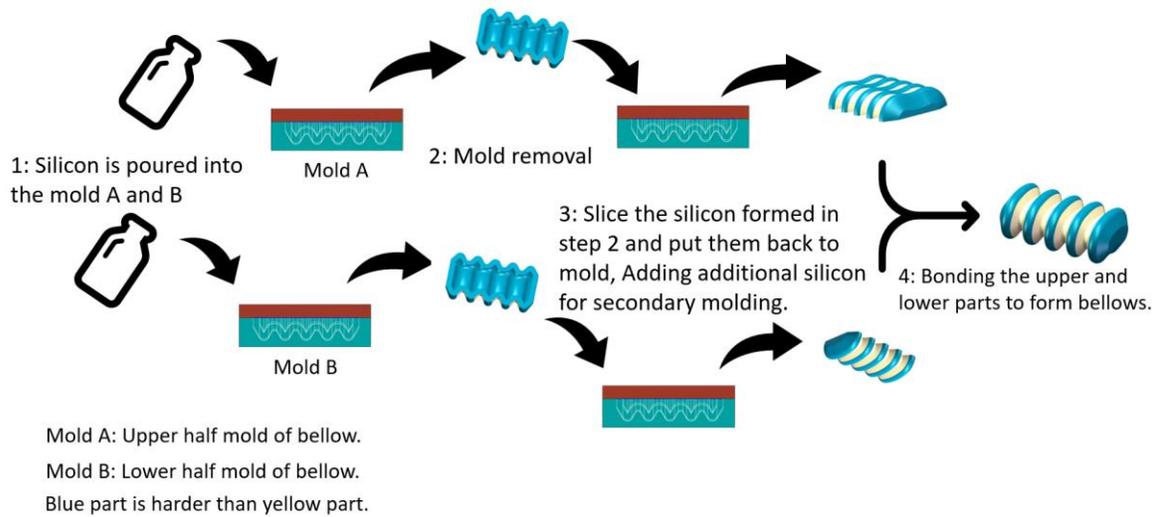

Fig. 3. Fabrication of the soft-bodied bellows.

In order to limit bellows' radial deformation and achieve maximum axial deformation, two different silicon hardness materials were used for its manufacturing: the shore hardness is equal to 30A of the blue part (Mould-star, Smooth-On, USA), and the shore hardness A is 00-30 of the yellow part (Eco-flex 00-30, Smooth-On, USA) (shown in Figure 3). An air pump has been used to remove the air from the silicon mixture in order to prevent bubbles from forming during curing, which can enhance the strength of the silicon and prevent fractures in the cavity.

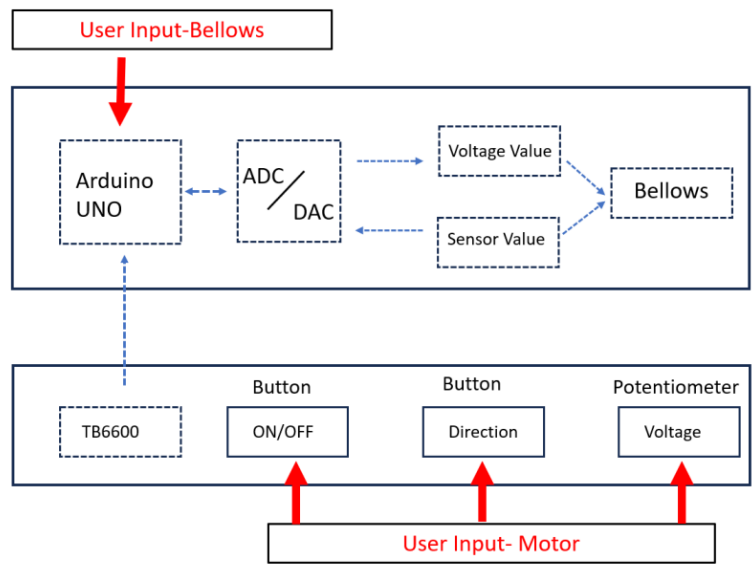

Fig 4. Open loop control of four independent bellows using Arduino microcontroller.

The pressure of the four independent bellows and the motor's angular speed are set by the user. Red characters in Fig. 4 indicate the user inputs, and blue dotted arrows indicate component data communication in the system.

## 3    Experiment setup

### 3.1 Force and displacement measurements of bellows

Fig. 5a shows three forces: Force A, Force B, force C (abbreviated as $F_A, F_B, F_C,$) at different points of the expansion mechanism when the bellows air pressure changes. $F_A$ represents the normal force at the top of the strut. $F_B$ denotes the radial force at the top of the strut, and $F_C$ is the axial force exerted by the bellow. The maximum normal bellow force, $F_A$ is 3.89N. The bellows operating pressure range is set to -101 to 283mbar as shown in Fig.6a. In order to evaluate the accuracy and consistency of the experimental results, 10 repeated measurements were made for each experiment without changing any experimental conditions. Each measurement is carried out independently, so as to evaluate the random error of the experiment and calculate the average and standard deviation of the data. This method allows us to obtain more reliable measurement results and better understand the random fluctuation of the experimental system.

The experimental results of the bellows displacement measurements were obtained using the laser sensor (optoNCDT 1420, MICRO EPSILON, Germany) shown in Fig. 5b. The bellows maximum lengthwise displacement is 8.98 mm. The maximum changes in other directions are also worthy of reference, which are 6.3mm in the height direction and 2.12mm in the width direction, respectively (refer to Fig. 6d). These results show that two silicon materials of different hardness can make a flexible linear actuator with relatively limited deformations in radial directions.

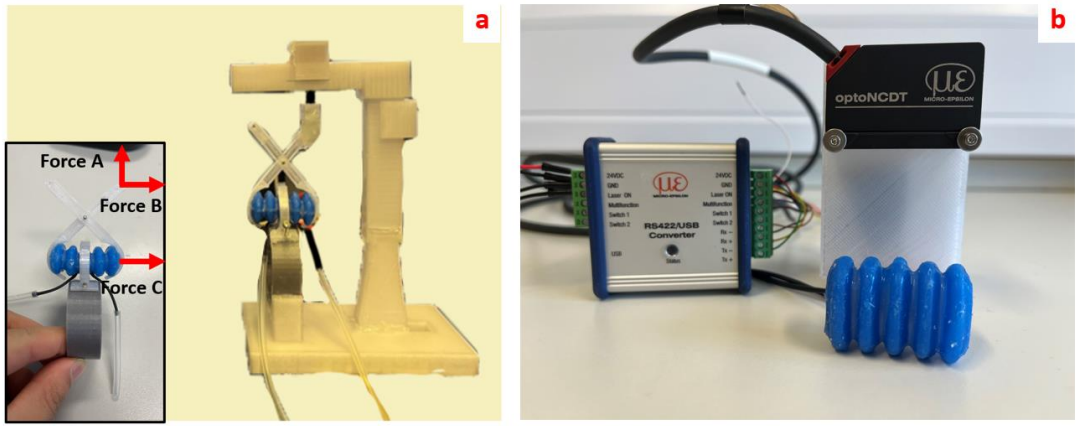

Fig. 5 a) Force sensor experimental platform of the bellow with strut (insert shows the direction and position of the force measured in the experiment); b) Laser sensor platform for detecting the deformation of bellows.

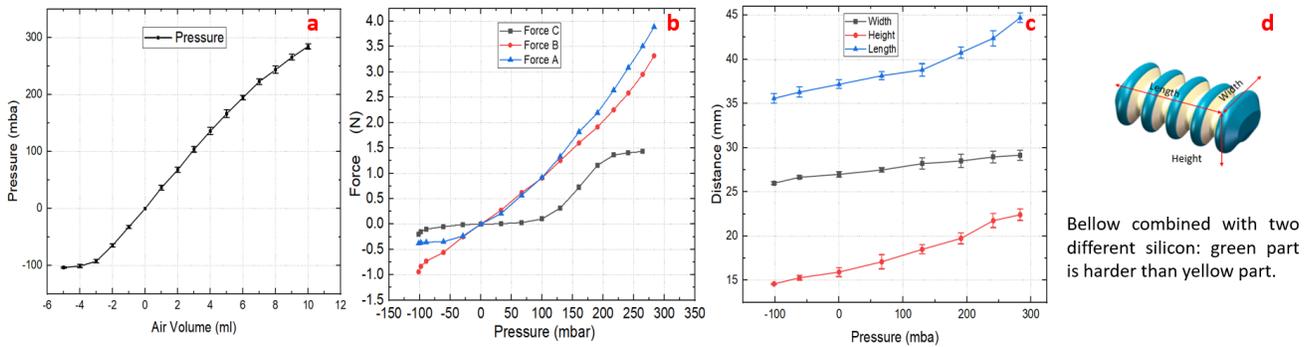

Fig. 6. a) The relationship between the air volume inside the bellows and the internal pressure; b) $F_A$, $F_B$ and $F_C$ as a function of pressure change in one bellow; c) Displacements of the bellow along width, height and length pressure in the bellow; d) Bellow rendering.

The four groups of expansion mechanisms are independently controlled by adjusting the air pressure in the bellows to adapt to the overall size and shape of the colon. According to Figure 6a, the prototype is designed at four times the intended final size (4:1), with a diameter 14 cm in its non-pressurized state. The experimental data show the average value of ten repeated measurements. Subject to the activation of the expansion mechanism, the diameter of the prototype changes by almost 53%, from 9.96 to 15.16 cm. When the bellows are subjected to negative pressure, they contract, causing the track frame diameter to expand beyond its non-pressurized state. Conversely, when the bellows are inflated with positive pressure, the expansion mechanism contracts, as shown in Figure 7.

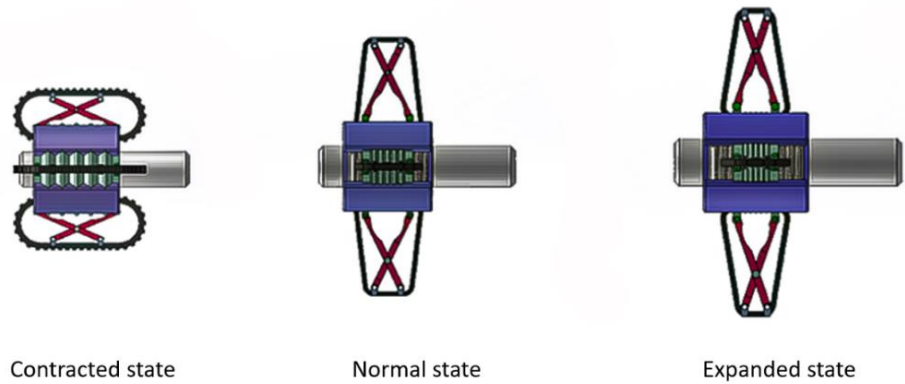

Fig. 7. Schematic diagram of working principle of expansion mechanism. Left - bellows are subjected to negative pressure; Middle - bellows are at atmospheric pressure, Right - bellows are subjected to positive pressure.

### 3.2 Prototype testing

#### 3.2.1 Testbed

A straightforward experimental platform was established to evaluate the prototype's performance. The testbed comprises several key components: a 3D-printed pipe support made from PLA, a transparent acrylic pipe, a force sensor, and the prototype system.

The test bed for this experiment is in the form of a rigid acrylic tube that simulates a straight colon section. This was used to test the propulsion and expansion capabilities of the prototype. This does not capture the flexible and tortuous nature of the human colon but provides valuable data to verify the feasibility of the concept.

The functionality of the prototype has been tested on a corresponding-size testbed. To explore different traction forces, we used three different surfaces, including a commercial bowel artificial tissue sheet (Bowel-9mm, Lifelike, Canada) and a transparent acrylic tube to simulate the colon. A 3D-printed bracket was used to secure the tube and prevent deviation during testing. Furthermore, Additionally, we varied the friction coefficient of the tube's inner surface by using both tissue and foam.

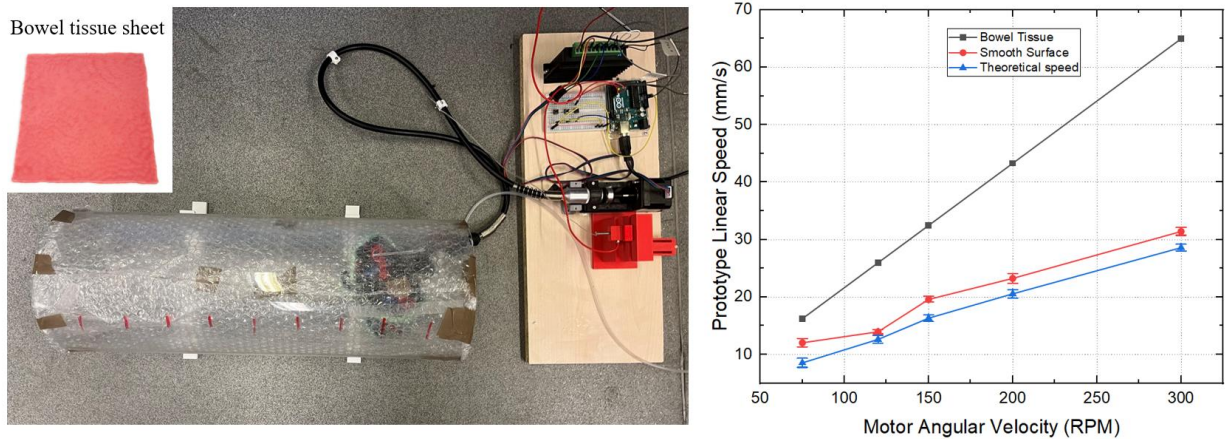

Fig. 8. Locomotion experiment of the prototype. The motor angular velocity was set to 75, 120, 150, 200 and 300 RPM to explore corresponding linear velocity of the prototype; Different friction coefficients of the tube surface were emulated using foam, artificial tissue (with mucus) and smooth surface.

Figure 8 compares the theoretical and tested linear velocities of the prototype on different surfaces. Due to friction losses between the flexible shaft, worm gear, tracks, and the tube surface, the actual velocities are lower than the theoretical values that do not account for friction. The experimental data represent the average of five repeated measurements. Although the prototype can achieve maximum speeds of 32.04 mm/s on a smooth surface and 29.29 mm/s on wet tissue at the maximum motor speed of 300 RPM, this speed may be impractical for operation inside the human colon. However, it is important to note that these figures are meant to demonstrate the prototype's maximum capability in its oversized version. In practical application, the motor speed can be precisely controlled to achieve a more appropriate and safe movement speed within the colon, ensuring effective operation without compromising patient safety.

### 3.2.2 Prototype propelling force

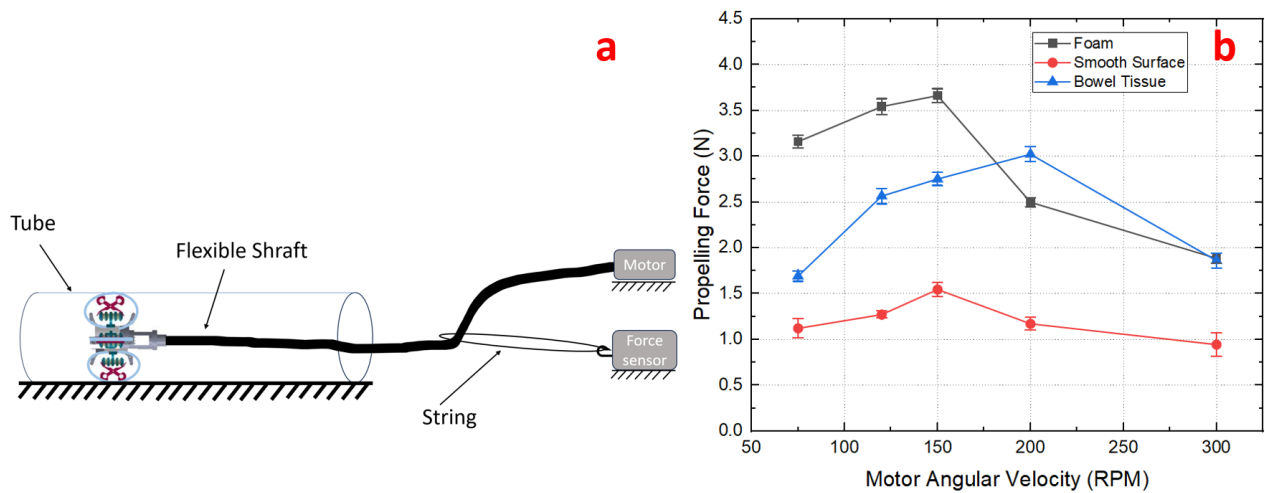

Fig. 9. a) Propelling force experiment schematic diagram; b) Comparison diagram of different propelling force and motor speed under three different friction coefficient surfaces.

The force sensor (LSb201, FUTEK, USA) is fixed on the platform and connected with the flexible shaft by string so as to obtain the propelling force of the prototype. With the three different friction coefficients of smooth tube, foam, and artificial bowel tissue, we tested the change in propelling force of the prototype by varying the motor speed in the range of 75-150 RPM, and passed five groups of repeated experiments, respectively, as shown in figure 9b. In the range of 75–150 RPM, the force changes as predicted, but with a further increase in speed, the propelling force of the prototype decreases. This can be explained by taking into account that at the motor's high speed, the transmission efficiency of the flexible shaft starts to decline, which consequently affects the worm and track coupling and leads to slippage. The experiment shows that the maximum propelling forces that can be achieved are 1.47, 2.83, and 3.61 N on smooth surfaces, artificial bowel tissue, and foam surfaces, respectively. It is worth noting that in the actual testing process, the movement of the device on the tissue is more stable. The potential reason may be that this material is softer. It is due to a soft contact surface that is more likely to produce a concave-like deformation, which increases its contact area. But specific reasons need to be further explored.

### 3.2.3 Experiment of diameter size-adapting

In this experiment, four tube sizes (10, 12, 14, and 15 cm) were used to verify the shape adaptability of the prototype. According to the previous results, the prototype gets maximum propulsion when the angular speed of the motor is 150–200 RPM (Fig. 9 b). Adhering to the principle of controlling variables, the motor rotation speed was fixed at 200 RPM. The inner surface of the large plastic tube was lined with artificial tissue to simulate a realistic colon surface. Under these controlled parameters, the variation in propulsion force of the prototype was measured as it navigated different pipe diameters, as depicted in Figure 10.

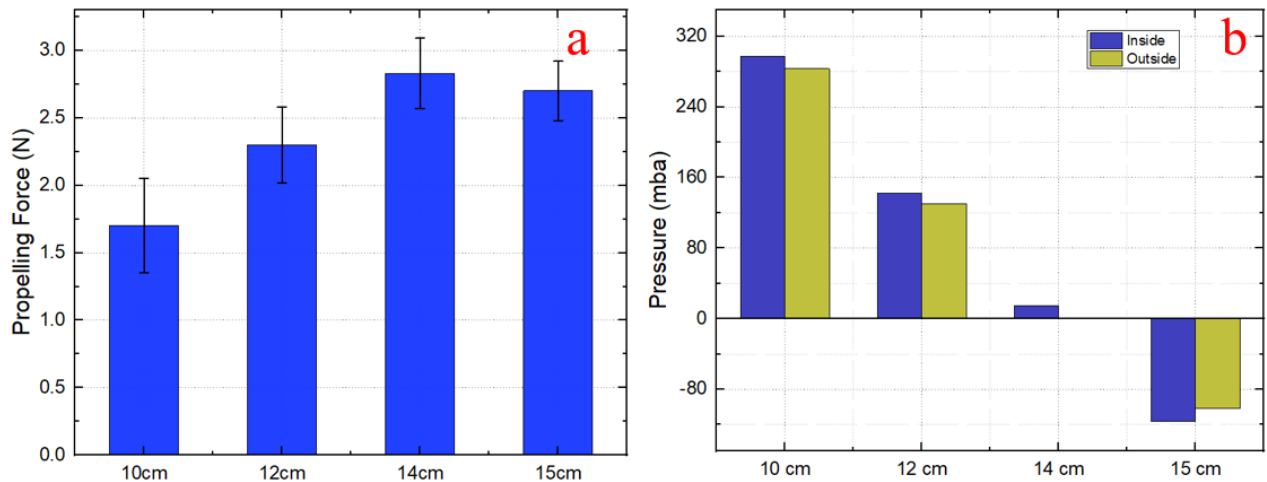

Fig. 10. Effect of pipe diameter on bellows pressure and propelling force in a prototype expansion mechanism. a ) Expansion mechanism in normal state; b) Conditioned maintenance of stable traction force.

As the prototype traverses a smaller tube, the traction force decreases with the reduction in pipe diameter while maintaining constant air pressure in the bellows. This is due to the increase in normal force between the pipe wall and the expandable tip. At this point, the struts are exposed to this higher normal force which ultimately affects the bellows. Since the bellows and struts are critical for the expansion, we explored how the pressure in bellows changes in the four different pipe diameters.

The blue bars labelled 'inside' in Fig. 10b indicate the average pressure that the bellows must maintain to achieve consistent propulsion force. Conversely, the yellow bars show the required theoretical pressure of the prototype without any external constraints, labelled outside." Blue and yellow bars show a significant difference in air pressure between these two conditions.

This discrepancy occurs when the expanding tip has to overcome its own weight as well as the stress exerted by the pipe's normal forces when the pipe's diameter is smaller than the expansion tip's normal state diameter. Consequently, the expansion mechanism must supply extra air pressure to counterbalance these forces and in order to maintain the traction. In its normal state, the diameter of the expandable tip is 14 cm, with the bellows' air pressure at ambient levels represented as 0 in Fig. 10b. Even in this state, there is a positive pressure in the bellows that counteracts the expandable tip's own weight.

## 4   Discussion

In this paper, we experimentally determined the propelling force of the prototype under various surface roughness conditions but did not measure the traction force. Propelling force drives the entire prototype, while traction force is exerted by end-effectors (wheels, tracks). In this paper, the traction force is influenced by the friction coefficient between the tracks and the contact surface, as well as the normal stress on the tracks. Research indicates that a traction force of at least 1 N is required to reliably drive the robot [19]. Each wheel of the Rollerball robotic endoscope generates about 0.33 N [13], and the SoftSCREEN achieves up to 2 N [15]. In our experiments, the maximum propelling force obtained on the artificial bowel tissue was 3 N. But propelling force and traction force are different in a strict sense, which means that the traction force of our prototype may not be higher compared to other models. Therefore, in the next step, we plan to study the relationship between traction force and propelling force by adjusting the friction coefficient of the track and increasing the normal stress on the track. Besides, the excessive force exerted on colon will cause mucosal damage. Therefore, it is also of great significance to carry out experimental research on traction force to evaluate the safety of the device.

Previously, we noted that the diameter of the colon variability is approximately 73%. The expansion rate of our prototype is about 53%, while SoftSCREEN prototype can extend almost 70% from its normal diameter. Generally speaking, smaller prototype size has higher expansion rate at the same elongation. In our prototype, the expansion rate is primarily determined by the length of the bellow and the struts. Although, it remains uncertain whether the size of prototype correlates with traction force, optimizing the structure of expansion mechanism and miniaturizing the prototype are potential solutions to improve the expansion rate.

In order to simplify the prototype analysis, a rigid testbed was used that does not emulate flexible colon walls. This allowed us to focus on the internal mechanism variables and less so on the interaction with the deformable environment. This setting admittedly poses limitations since the flexible colon wall would affect the contact area with the tracks and therefore the traction force of the mechanism.

In the propulsion testing experiment, from the perspective of operator, the tip the prototype has a more stable movement compared to other friction coefficient surfaces, and from the results, the setting of 200 RPM in Figure 9 b) obtains higher propulsion force. The potential reason is that the soft surface, because of its greater deformation trend, enhances the traction and is conducive to the smoother movement of the prototype. We need to do more in-depth experiments to reveal the relationship.

## 5   Conclusion and Future work

In this paper, a design of a shape-fitting robot endoscope with an expansion mechanism is proposed with the purpose of achieving consistent traction across different colon diameters. The expansion mechanism of the prototype consists of four groups of bellows and struts. The four groups can be independently controlled through the pressures in the bellows, which could significantly enhance the prototype's flexibility, allowing for more complex asymmetric changes and functions, such as steering around the tortuous colon bends. This aspect is not explored in this paper and will be addressed in future work. The experimental results verify the effectiveness of the design in the simulated colon environment, showing good propulsion performance and diameter size adaptability.

The future work is not limited to optimizing the expansion mechanism, but also needs to further improve the expansion rate of the robot. Specifically, it is necessary to explore the influence of flexible materials on the robot's movement efficiency, such as whether the track can produce greater traction in soft artificial tissues, measure the traction of the track, and try to explain the relationship between it and traction. In addition, it is more important to test and evaluate the steering function of the robot, and we will verify this function by setting pipes with different angles. It is also the content of future research to realize the autonomous adjustment of robot function (such as expansion mechanism) by adding sensors in robot system.

In a word, the robot endoscope designed in this paper shows great potential in colonoscopy. In the future, through further clinical trials and technical optimization, this innovative design is expected to be widely used in practical medical applications, improve the efficiency and safety of colonoscopy, and provide strong support for the early detection and prevention of intestinal diseases.

**Author Contributions**

1. Jiayang Du: Conceptualization, methodology implementation, design and fabrication, experimentation and validation.
2. Sanja Dogramazi: Conceptualization, experimental instruction, supervision, manuscript editing.
3. Lin Cao: Conceptualization, experimental instruction, supervision, manuscript editing.

**Funding**

University of Sheffield Institutional Open Access Fund